\documentclass[10pt,journal,compsoc]{IEEEtran}
\usepackage{amsmath,amsfonts}
\usepackage{algorithmic}
\usepackage{algorithm}
\usepackage{array}
\usepackage{textcomp}
\usepackage{url}
\usepackage{verbatim}

\usepackage{subfiles}
\usepackage{booktabs}
\usepackage{multirow}
\usepackage{stfloats}
\usepackage{lipsum}
\usepackage{arydshln}
\usepackage{graphicx}
\usepackage{subcaption}
\usepackage[normalem]{ulem}

\usepackage{amssymb}
\usepackage{tikz}
\usepackage{amsmath}
\usepackage{enumitem}
\usepackage{xcolor}
\usepackage{soul}
\usepackage{ragged2e}
\usepackage{bm}

\newcommand{\vic}[1]{\textcolor{black}{#1}}

\ifCLASSOPTIONcompsoc
  \usepackage[nocompress]{cite}
\else
  \usepackage{cite}
\fi
\ifCLASSINFOpdf
\else
\fi
\begin{document}
%
\title{COTET: Cross-view Optimal Transport for Knowledge Graph Entity Typing}
%
%
%
%

\author{Zhiwei Hu, Víctor Gutiérrez-Basulto, Zhiliang Xiang, Ru Li, 
Jeff Z. Pan
\IEEEcompsocitemizethanks{\IEEEcompsocthanksitem Z. Hu and R. Li are with the School of Computer and Information, Shanxi University, Shanxi, China. 
E-mail: zhiweihu@whu.edu.cn; liru@sxu.edu.cn

\IEEEcompsocthanksitem V. Gutiérrez-Basulto and Z. Xiang are with the School of Computer Science and Informatics, Cardiff University, Cardiff, UK. 
E-mail: gutierrezbasultov@cardiff.ac.uk; xiangz6@cardiff.ac.uk

\IEEEcompsocthanksitem J. Pan is with the school of Informatics, University of Edinburgh, Edinburgh, UK. 
E-mail: j.z.pan@ed.ac.uk
}

}

\IEEEtitleabstractindextext{%

\justify \begin{abstract}
\vic{Knowledge graph entity typing (KGET) aims to infer missing entity type instances in knowledge graphs. Previous research has predominantly centered around leveraging contextual information associated with entities, which provides valuable clues for inference. However, they have long ignored the dual nature of information inherent in entities, encompassing both high-level coarse-grained cluster knowledge and fine-grained type knowledge. This paper introduces  \textbf{C}ross-view \textbf{O}ptimal \textbf{T}ransport for knowledge graph \textbf{E}ntity \textbf{T}yping (\textbf{COTET}), a method that effectively incorporates the information on how types are clustered into the representation  of entities and types. COTET comprises three modules: i) \textit{Multi-view Generation and Encoder}, which captures structured knowledge at different levels of granularity through \textit{entity-type}, \textit{entity-cluster}, and \textit{type-cluster-type} perspectives; ii) \textit{Cross-view Optimal Transport}, transporting view-specific embeddings to a unified space by minimizing the Wasserstein distance from a distributional alignment perspective; iii) \textit{Pooling-based Entity Typing Prediction}, employing a mixture pooling mechanism to aggregate prediction scores from diverse neighbors of an entity. Additionally, we introduce a distribution-based loss function to mitigate  the occurrence of false negatives  during training. Extensive experiments demonstrate the effectiveness of COTET when compared to existing baselines.}

\end{abstract}

}

\maketitle

\section{Introduction}
\begin{figure}[t!]
    \centering
    \includegraphics[width=0.46\textwidth]{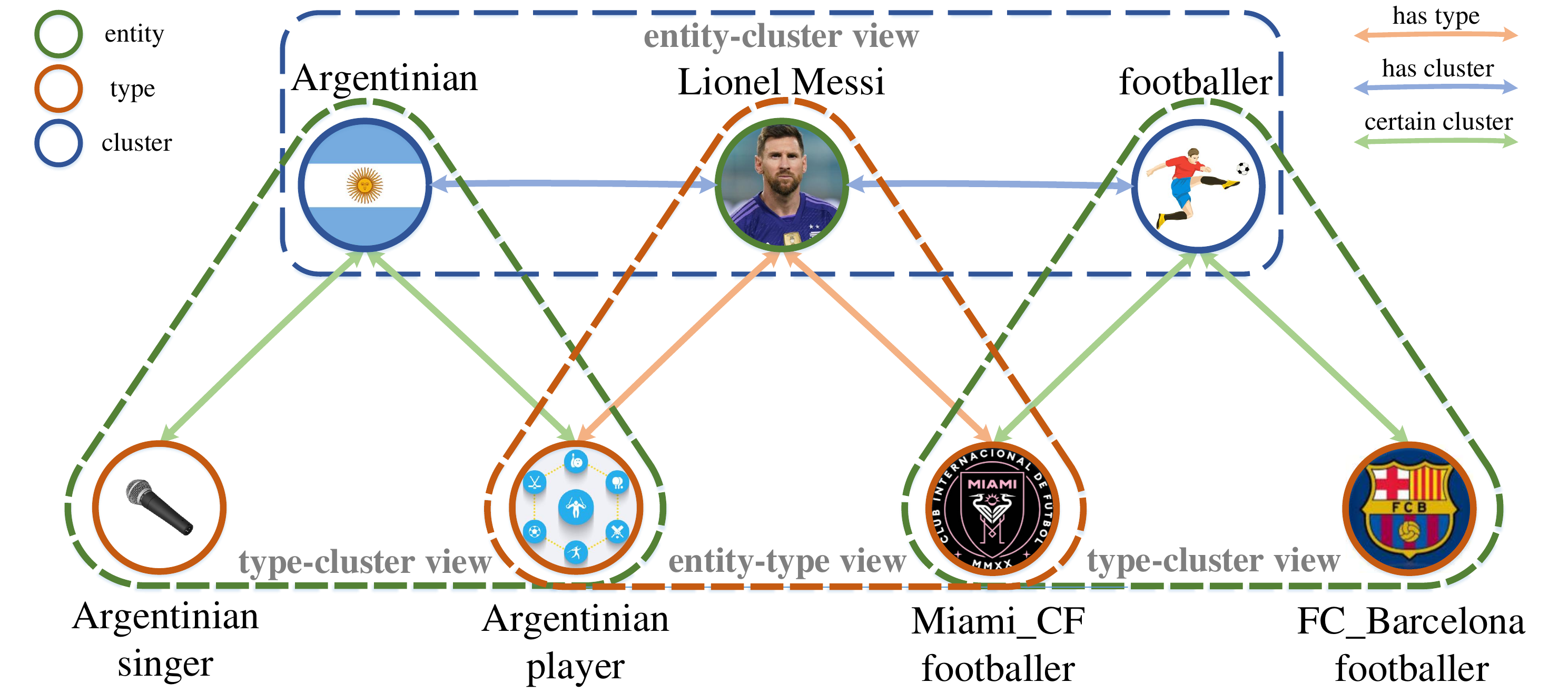}
    \caption{A KG from three perspectives: \emph{entity-type view}, \emph{entity-cluster view}, and \emph{type-cluster view}. 
    }
    \label{figure_instance}
\end{figure}

\IEEEPARstart{K}{nowledge} graphs (KGs) represent factual knowledge using  triples of the form $(e, r, f)$, capturing that  entities  $e$ and $f$ are connected via the  relation type $r$. KGs also contain  entity type assertions of  the form $(e, \textit{has\_type}, t)$, denoting  that the entity $e$ has type $t$.  For example,  as show in Figure~\ref{figure_instance}, the entity \textit{Lionel Messi} is of types \textit{Argentinian\_player} and \textit{Miami\_CF\_footballer}. Entity type information is crucial in  many applications, including knowledge graph completion~\cite{Zijie_2023, Miao_2022, Guanglin_2022}, question answering~\cite{Zhiwei_2022_2, Yu_2019}, and entity alignment~\cite{Qian_2023, Zhenxi_2022}. 
Nevertheless, notwithstanding the rich type knowledge present in existing Knowledge Graphs (KGs), such as FB15k~\cite{Antoine_2013} and YAGO43k~\cite{Moon_2017}, their coverage remains significantly incomplete.
For example, in Figure~\ref{figure_instance}, the entity \textit{Lionel Messi} should also be of type \textit{FC\_Barcelona\_footballer}, but in reality this content is missing. \vic{For instance, in the  FB15k dataset, 10\% of entities have the type \textit{/music/artist}, but are missing the type type \textit{/people/person}~\cite{Moon_2017}.}

\vic{To address this problem, we focus on the \textit{Knowledge Graph Entity Typing} (KGET) task, which aims to infer  missing  entity type assertions in KGs.}
Several efforts have been devoted to handling the KGET task, including embedding-based~\cite{Antoine_2013, Trouillon_2016, Zhiqing_2019, Xiou_2023, Liang_2022, Moon_2017, Zhao_2020, Ge_2021}, graph neural networks (GNNs)-based~\cite{Jin_2019, Zhuo_2022, Zhao_2022, Zou_2022, Vashishth_2020, Michael_2018, Pan_2021, Jin_2022}, and transformer-based methods~\cite{Wang_2019, Chen_2021, Hu_2022}. \vic{However, each of this kind of methods has its own drawbacks. Embedding-based methods encode all neighbors of target entity into a single vector, but in many cases only some neighbors are necessary to infer the correct types.
GNNs-based frameworks use expressive representations for entities and relations based on their neighbors, however they mainly aggregate information  along the paths starting from neighbors of the target entity, falling short on capturing the interactions between non-directly connected neighbors~\cite{Hu_2022}. Transformer-based methods  rely on computationally heavy Transformer~\cite{Ashish_2017} structures to encode entities and their relations, as well as known type neighbors to infer missing types. More importantly, all existing methods \textit{only consider a single  view of knowledge}. However,  in practice, type information in KGs   can be categorized.}
For example, in Figure~\ref{figure_instance}, the types \textit{Miami\_CF\_footballer} and \textit{FC\_Barcelona\_footballer} can be clustered together in the class \textit{footballer}, and  \vic{the types \textit{Argentinian singer} and \textit{Argentinian player} can be clustered into the class \textit{Argentinian}}.  The introduction of such cluster knowledge will help to refine the original \textit{entity-type view} by generating different perspectives of knowledge, namely, \textit{entity-cluster view} and \textit{type-cluster view}.

\vic{Modeling three-views for KGET is non-trivial, where the challenges are two-fold: On the one hand, for the same entity, type or cluster,  embeddings  with different perspectives will be obtained from different views, so we need some strategy to bridge the knowledge from these different views. For example,  for entity embeddings, effective semantic mappings from the fine-grained entity-type view to their corresponding coarse-grained entity-cluster view need to be carefully designed. In this case, the main question is  ``\textit{how to effectively combine embeddings obtained from heterogeneous view spaces}''. On the other hand, after introducing  cluster information, to predict the types of an entity we can utilize the information provided by its  neighbors. Different neighbors will make their respective predictions, and then we need to adequately combine these prediction scores to form the final result. The main issue here is  ``\textit{how to effectively aggregate the inference results from different neighbors}''}.


To address these points,  we propose   \textbf{COTET}, a \textbf{C}ross-view \textbf{O}ptimal \textbf{T}ransport model for  knowledge graph \textbf{E}ntity \textbf{T}yping. COTET  contains three modules:  \textit{multi-view generation and encoder}, \textit{cross-view optimal transport}, and  \textit{pooling-based entity typing prediction}. The multi-view generation and encoder module aims to generate the entity cluster and type cluster views,  after introducing the cluster information. Then it uses different GNN encoders to obtain the node embeddings for each view. We further introduce a cross-view optimal transport module to generate a consistent representation in a unified space from different heterogeous view spaces. To this end, the fusion of different view embeddings of the same entity, type, or cluster is recasted as an optimal transport problem: we move different view embeddings to a unified aligned space by minimizing the Wasserstein distance between different distributions. To effectively aggregate the inference results from different neighbors, we  introduce a pooling-based entity typing prediction module, which employs a  pooling method, including multi-head weight pooling, max pooling, and average pooling. In addition, we propose a Beta probability distribution based cross-entropy loss to mitigate the  false-negatives  during training.

\vic{
Our contributions are summarized as follows:
\begin{itemize}[itemsep=0.5ex, leftmargin=5mm]
\item We propose a  method which encodes knowledge from different perspectives and at different levels of granularity.
\item We introduce a  cross-view optimal transport module to align the embeddings from different views, and a mixture pooling strategy to aggregate prediction scores from different neighbors.
\item We design a distribution-based loss function to alleviate the false-negative problem during training. Further we conduct thorough experiments with a series of ablation studies on two real-world knowledge graphs.
\end{itemize}}

\section{Related Work}
\subsection{\vic{Embedding-based Methods}}
After introducing the relation type \textit{has\_type} to connect entities and types, the KGET can be formulated as a knowledge graph completion (KGC) task. \vic{Therefore,  traditional KGC methods such as TransE~\cite{Antoine_2013}, ComplEx~\cite{Trouillon_2016}, RotatE~\cite{Xiou_2023}, CompoundE~\cite{Zhiqing_2019} and SimKGC~\cite{Liang_2022} can be seamlessly migrated to the KGET task. In addition, some embedding-based models customized to the KGET task have also proposed.} ETE~\cite{Moon_2017} follows the TransE~\cite{Antoine_2013} approach to obtain the entity and relation embeddings, and then builds a mechanism to shorten the distance between entity and type embeddings. ConnectE~\cite{Zhao_2020} embeds entities and types into different spaces, and then uses the E2T mechanism to map entities from  the entity space into  the type space. CORE~\cite{Ge_2021}  leverages RotatE~\cite{Zhiqing_2019} and ComplEx~\cite{Trouillon_2016} to get entity and type embeddings, and uses a regression model to capture the relatedness between them. \vic{Despite their simplicity and intuitiveness, these methods ignore the rich  information provided by the neighbors of an entity, which seriously limits their performance.}


\subsection{\vic{Graph Neural Network-based Methods}}
\vic{Since Graph Neural Networks (GNNs) are inherently capable of encoding the structure of the neighborhood of an  entity, some methods commonly used in KGC, such as HMGCN~\cite{Jin_2019}, CompGCN~\cite{Vashishth_2020}, and RGCN~\cite{Michael_2018}, can be easily migrated to the KGET task. In addition, many bespoke  models have  been proposed, in which  there is only one relation between entities and types: \textit{has\_type}).} AttEt~\cite{Zhuo_2022} develops a type-specific attention mechanism to aggregate neighborhood knowledge with the corresponding entity to match the candidate types. \vic{ConnectE-MRGAT~\cite{Zhao_2022} builds a novel heterogeneous relational graph (HRG), and proposes a multiplex relational graph attention network to learn on HRG, then utilizes a connecting embeddins model to make entity type inference. RACE2T~\cite{Zou_2022} proposes an encoder FRGAT and decoder CE2T, where FRGAT uses the scoring function of KGC methods to calculate the attention coefficient between entities, and CE2T performs entity type prediction.} CET~\cite{Pan_2021} utilizes  neighborhood information in an independent-based mechanism and aggregated-based mechanism for inferring missing entity types. MiNer~\cite{Jin_2022}  aggregates both one-hop and multi-hop neighbors, and further predicts the entity types by using a type-specific local inference and a type-agnostic global inference. Although these methods can indeed encode the entity's neighborhood information, they do not consider the coarse-grained cluster information related to entity types.


\subsection{\vic{Transformer-based Methods}}
\vic{Transformers, which measure the semantic similarity of neighbors, have become increasingly popular for addressing various KG-related tasks~\cite{Yang_2022, Hongcai_2023, Wang_2019, Chen_2021, Zhiwei_2023}. \vic{CoKE~\cite{Wang_2019} and HittER~\cite{Chen_2021}, which have good performance in the KGC field, can directly applied to KGET task.} In addition, TET~\cite{Hu_2022} introduces three transformer modules to encode local and global neighborhood information. It further uses  class membership of types to semantically strengthen the entity representation. However,  a main drawback is that transformer-based methods have a substantial time overhead as  an additional transformer module needs to be set up to ensure that the structural content of the graph is captured.}

\subsection{\vic{Optimal Transport}}
\vic{Optimal transport (OT) is a fundamental mathematical tool which aims to derive an optimal plan to transfer one distribution to another, OT provides an elegant framework to compare and align distributions~\cite{Arjovsky_2017, YenChun_2020, Wei_2023}. There are extensive studies utilizing the transport plan of OT to solve assignment problems in  computer vision~\cite{Nicolas_2011, Justin_2015}, domain adaption~\cite{Xiang_2022, Wanxing_2022}, and unsupervised learning~\cite{Mathilde_2020, Yuki_2020}. OTKGE~\cite{Zongsheng_2022} models the multi-modal fusion procedure as a transport plan moving different modal embeddings to a unified space by minimizing the Wasserstein distance between multi-modal distributions. OTEAE~\cite{Amir_2022} employs optimal transport to induce structures of documents based on sentence-level syntactic structures for the event argument extraction task. SCCS~\cite{Jielin_2023} follows a cross-domain alignment objective with optimal transport distance to leverage multimodal interaction to match and select the visual and textual summary. CMOT~\cite{Yan_2023} finds the alignment between speech and text sequences via optimal transport and then combines the sequences from different modalities at a token level using the alignment. However, existing studies lack an alignment mechanism between multi-view embedding representations. To the best of our knowledge, we are the first to adopt the optimal transport mechanism for KGET task.}

\section{Background}
\smallskip \noindent \textbf{Task Definition.} 
\vic{Let $\mathcal E$, $\mathcal R$ and $\mathcal {T}_p$ respectively be finite sets of \emph{entities}, \emph{relation types} and \emph{entity types}. We consider  \emph{knowledge graphs $\mathcal{K}$} composed of a \emph{relational graph} $\mathcal{G}$ and a \emph{type graph} $\mathcal T$.  $\mathcal G$ and $\mathcal T$ are respectively composed by a set of triples of the form $(e,r,f)$ and $(e,\emph{has\_type}, t)$, with $e,f \in \mathcal E$, $r \in \mathcal R$, $t \in \mathcal T_p$, and \emph{has\_type}  a special relation type (not occurring in $\mathcal R$) used to connect entities with their corresponding types. Crucial to inferring missing types is the information provided by the neighbors of an entity.  For an entity $e$, we define its \textit{relational neighbors} $\mathcal{N}_e^\mathcal G = \{(r,f) \mid (e,r,f) \in \mathcal{G}\}$ and its \textit{type neighbors} $\mathcal{N}_e^\mathcal T = \{(\emph{has\_type},t) \mid (e,\emph{has\_type},t) \in \mathcal{T}\}$. In this paper, we study the \textit{Knowledge Graph Entity Typing (KGET)} task which aims at inferring missing types from $\mathcal T$ of entities, and thus at completing the type sub-graph $\mathcal T$ of $\mathcal K$. 
}


\smallskip \noindent \textbf{Type Clustering.} \vic{Just like ``birds of a feather flock together'', entity types often tend to bunch together in some way. For instance, the types \textit{football\_player}, \textit{basketball\_player} and \textit{baseball\_player} all belong to the category  \textit{player}.  Entities typically have associated a hierarchical label, including both coarse-grained cluster and fine-grained type information. The coarse-grained cluster content can be used to narrow down the scope of fine-grained type labels, thereby restricting the decision-making space for inferring missing type knowledge. Keeping this in mind, we  introduce cluster information into the type prediction process. \vic{For the FB15kET dataset, we can directly obtain cluster information using a rule-like approach based on their hierarchical type annotations. For instance, the type \textit{/location/uk\_overseas\_territory} belongs to the cluster \textit{location} and the type \textit{/education/educational\_degree} belongs to the cluster \textit{education}. The source dataset YAGO of YAGO43kET provides an alignment between types and WordNet concepts\footnote{https://yago-knowledge.org/downloads/yago-3}. So, we can directly obtain the words in WordNet describing the cluster to which a type belongs to. For example, for the type \textit{wikicat\_Football\_clubs\_in\_Ghana}, its cluster is \textit{wordnet\_club\_108227214}, and for the type \textit{wikicategory\_Male\_actors\_from\_Arizona}, its cluster is \textit{wordnet\_actor\_109765278}.}}


\section{Method}\label{sec:method}
\begin{figure*}[t!]
    \centering
    \includegraphics[width=1.0\textwidth]{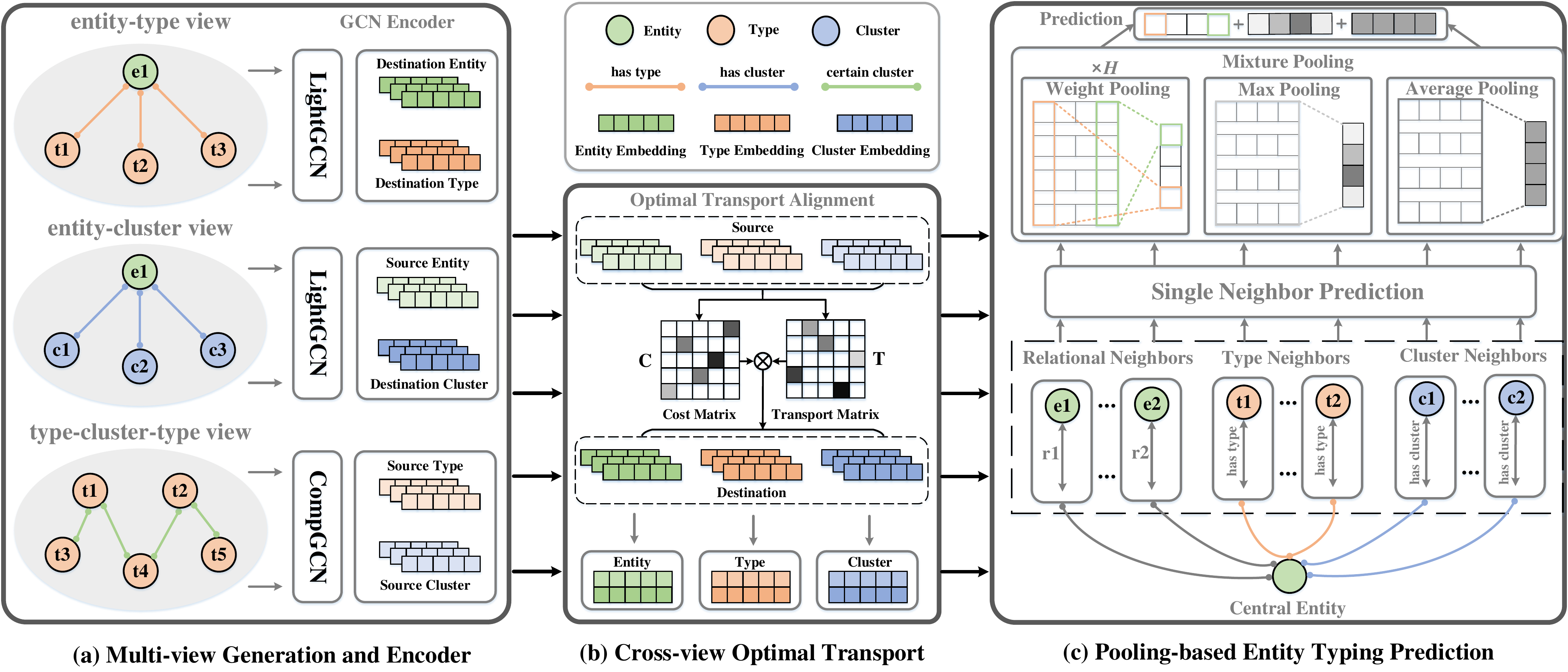}
    \caption{An overview of our COTET model, containing three modules: (a): Multi-view Generation and Encoder, (b): Cross-view Optimal Transport, and (c): Pooling-based Entity Typing Prediction.}
    \label{figure_model}
\end{figure*}

In this section, we present the three main components of COTET. First, we show how  different views are generated and encoded. Then, we discuss how to use the optimal transport mechanism to align view-specific embeddings  into a unified space. Finally, we present a mechanism to aggregate the prediction results from different neighbors.

\subsection{Multi-view Generation and Encoder}
 \vic{In our method, COTET, we use both relational neighbors and type neighbors for inferring missing entity types. Further, we leverage the   hierarchical knowledge of types inherently available in KGs, providing  coarse-grained cluster and fine-grained type information. Specifically, after introducing cluster knowledge into the type graph $\mathcal T$,  we construct a heterogeneous graph with \textit{entity-type}, \textit{entity-cluster}, and \textit{type-cluster}  connections. To fully exploit the different levels of abstraction provided by these connections, we split the heterogeneous graph into three graphs: \textit{entity-type view}, \textit{entity-cluster view}, and \textit{type-cluster-type view}.} 



\smallskip \noindent \textbf{Entity-Type View.} \vic{The entity-type view, denoted as $\mathcal{T}_{e2t}$, is given by the original type graph $\mathcal T$. We will leverage the type knowledge in $\mathcal{T}_{e2t}$ for inferring missing types. For example, given the type assertion (\textit{Lionel Messi}, \textit{has\_type}, \textit{Argentinian\_footballers}), we could deduce the missing type assertion (\textit{Lionel Messi}, \textit{has\_type}, \textit{footballers}), since the type \textit{Argentinian\_footballers} entails \textit{footballers}. }


\smallskip \noindent \textbf{Entity-Cluster View.} \vic{The entity-cluster view, denoted as $\mathcal{T}_{e2c}$, is  generated from the entity-type graph $\mathcal{T}_{e2t}$ and the clusters to which  types belong to. The entity-cluster view provides a coarser perspective than the entity-type view by allowing to relate entities to the classes to which their types belong to.  It also helps establishing the decision boundaries between different types. For instance, given an assertion ($e, \textit{has\_type}, t)$ and a cluster $c$ to which $t$ belongs to, we can generate  a triple $(e, \textit{has\_cluster}, c)$ in $\mathcal{T}_{e2c}$. As an example, if the entity \textit{Lionel Messi} has type \textit{Argentinian\_footballers}, we can obtain the $\mathcal{T}_{e2c}$ triples (\textit{Lionel Messi}, \textit{has\_cluster}, \textit{Argentinian}) and (\textit{Lionel Messi}, \textit{has\_cluster}, \textit{footballers}), as the type \textit{Argentinian\_footballers} belongs to the clusters \textit{Argentinian} and \textit{footballers}. }


\smallskip \noindent \textbf{Type-Cluster-Type View.} 
\vic{The type-cluster-type view, denoted as $\mathcal{T}_{tct}$, is obtained from the clusters and the types they contain. In principle,  to construct $\mathcal{T}_{tct}$ we could simply connect types to their corresponding cluster using a special relation type \textit{belongs\_to}. However, this construction  does not fully capture the semantic information provided by the relation between types and their clusters, as a type often belongs to multiple clusters and a cluster typically contains multiple types. Instead,   to construct $\mathcal{T}_{tct}$ we  use a cluster name as a relation type to connect all types that belong to that cluster, effectively forming  a multi-relational graph.  
For example, for the cluster \textit{player}, which includes the types \textit{football\_player}, \textit{basketball\_player}, \textit{American\_player}, and \textit{Argentinian\_player}, we can add the following triples to $\mathcal{T}_{tct}$: (\textit{football\_player}, \textit{player}, \textit{basketball\_player}) and (\textit{American\_player}, \textit{player}, \textit{Argentinian\_player}). Note that even if these two triples contain the same relation type \textit{player}, the domains of the entities to which they connect are different. For example, \textit{football\_player} and \textit{basketball\_player} are related to ball sports, while \textit{American\_player} and \textit{Argentinian\_player} are related to nationalities. Thus, by using cluster names as relations, we can capture multi-faceted information between types. 
}

\smallskip \noindent \textbf{Multi-View Encoder.}
\vic{We employ graph convolutional networks (GCNs)~\cite{Thomas_2017} to encode different views and obtain embedding representations of entities, types, and clusters from different perspectives.  More precisely,  we utilize LightGCN~\cite{Xiangnan_2020} to encode  $\mathcal{T}_{e2t}$ and $\mathcal{T}_{e2c}$  because they are single-relational graphs.  However, for $\mathcal{T}_{tct}$,   given its multi-relational structure, we use CompGCN~\cite{Vashishth_2020} as the graph encoder.} 

\smallskip
\vic{The embeddings   of an entity, a type or a cluster from different views usually occur  in different heterogeneous spaces. Thus, a direct fusion of these view-dependent embeddings   might destroy the intrinsic distribution and lead to an inconsistent  representation in the unified space. The optimal transport mechanism~\cite{Liqun_2020, Zongsheng_2022, Luc_2022, Jianheng_2023} can naturally transport different view embeddings into a unified space, while overcoming space heterogeneity by minimizing the Wasserstein distance between different distributions. The optimal transport mechanism contains two different roles: source and destination. The primary objective is to minimize the disparity between the source embeddings and the destination embeddings, ensuring their alignment in a shared space.}
%
\vic{With this in mind, we distinguish the different embeddings for the same target  (entity, type or cluster)  depending on the  view. We  respectively denote the entity and type embeddings obtained from $\mathcal{T}_{e2t}$ as $\textbf{E}_{e \leftarrow e2t}^{d}$ and $\textbf{E}_{t \leftarrow e2t}^{d}$, the entity and cluster embeddings obtained from $\mathcal{T}_{e2c}$ as $\textbf{E}_{e \leftarrow e2c}^{s}$ and $\textbf{E}_{c \leftarrow e2c}^{d}$, and the type and cluster embeddings obtained from $\mathcal{T}_{tct}$ as $\textbf{E}_{t \leftarrow tct}^{s}$ and $\textbf{E}_{c \leftarrow tct}^{s}$, where superscript $s$ and $d$ represent the source and destination, respectively. 
The choice of the source or destination roles can be explained as follows. Since  the entity embeddings obtained from the entity-type view are  `finer' than those from the entity-cluster view, we consider the former kind of embeddings as  destination ones and the latter as  source ones. Given the relatively small scale of the type-cluster-type view, the structure and semantic knowledge it can capture is relatively limited, so we consider the type and cluster embeddings obtained from the type-cluster-type view as sources. 
The entity-type and entity-cluster views are larger and denser than the type-cluster-type view, so we regard the type embeddings  and cluster embeddings  obtained from the entity-type  and entity-cluster views as  destination embeddings. In our experiments, we will show that if the roles of source and destination are reversed in the subsequent optimal transport mechanism, there is a significant impact on performance.}

\subsection{Cross-view Optimal Transport}
\smallskip \noindent \textbf{Wasserstein Distance.} \vic{The Wasserstein distance (WD) is commonly used for matching two distributions (\textit{e.g}., two sets of embeddings) ~\cite{Liqun_2020, Zongsheng_2022, Luc_2022, Jianheng_2023}. 
Let $\mathbb{X}=\{x_1, \ldots, x_n\}$ and $\mathbb{Y} = \{y_1, \ldots, y_m\}$ be two sets of  real numbers. Further, let $\bm{\mu} \in \textbf{P}(\mathbb{X})$  and $\bm{\nu} \in \textbf{P}(\mathbb{Y})$ be two discrete distributions, where $\bm{\mu}$ is a probability distribution defined on $\mathbb{X}$, and $\bm{\nu}$ is a probability distribution defined on $\mathbb{Y}$.  Intuitively, the Wasserstein distance will be defined as the minimum average transportation cost from $\bm{\mu}$ to $\bm{\nu}$.  Let  $\bm{\alpha}=(\alpha_1,\cdot\cdot\cdot,\alpha_n)$ and $\bm{\beta}=(\beta_1,\cdot\cdot\cdot,\beta_m)$ be weights that respectively belong to the $n$- and $m$-dimensional simplex, we formulate $\bm{\mu}$ and $\bm{\nu}$ as $\bm{\mu}=\sum_{i=1}^{n}\alpha_i\delta(x_i)$ and $\bm{\nu}=\sum_{j=1}^{m}\beta_j\delta(y_j)$, where $\delta(x_i)$ and $\delta(y_j)$ respectively denote the Dirac measure on $x_i$ and $y_j$. Note that  the Dirac measure can convert a discrete point into a continuous function.} 
%
%
%
 If $\bm{\mu}$ and $\bm{\nu}$ follow a uniform distribution, then  $\alpha_i=\frac{1}{n}$, and $\beta_j=\frac{1}{m}$. $\Pi(\bm{\mu}, \bm{\nu})$ denotes the set of all latent joint distributions with marginal distributions $\bm{\mu}$ and $\bm{\nu}$. The Wasserstein distance between the two probability distributions $\bm{\mu}$ and $\bm{\nu}$ is defined as: 
%
%
%
\begin{equation}
\mathcal{W(\bm{\mu}, \bm{\nu})}=\underset{\textbf{T} \in \Pi(\bm{\mu}, \bm{\nu})}{\textit{min}} \sum_{i=1}^{n}\sum_{j=1}^{m}\textbf{T}_{ij}\textbf{C}_{ij}
\label{equation_1}
\end{equation}
\vic{where $\Pi(\bm{\mu}, \bm{\nu})=\{\textbf{T} \in \mathbb{R}^{n\times m}|\textbf{T1}_m=\bm{\mu},\textbf{T}^{\top}\textbf{1}_n=\bm{\nu}\}$, \textbf{1} denotes an all-one vector, $\textbf{C}_{ij}$ is the transportation cost evaluating the distance between $x_i$ and $y_j$, (\textit{e.g}., the cosine distance between two embeddings), $\textbf{T}_{ij}$ represents the transport matrix, which measures the amount of mass shifted from $x_i$ to $y_j$. Intuitively, the transport plan \textbf{T} measures a probabilistic matching of values between two embeddings: the larger the value of $\textbf{T}_{ij}$ is, the more likely $x_i$ and $y_j$ are aligned}.


\smallskip \noindent \textbf{Optimal Transport Alignment.} \vic{Considering that different views focus on different information perspectives (thus obtaining embeddings with different levels of importance), to reduce the information discrepancy caused by view-specific constructions, we propose an \emph{optimal transport alignment (OTA) module}, which adopts an optimal transport approach~\cite{Zongsheng_2022,Jianheng_2023} to align the embeddings obtained from different views. Lets consider the embedding alignment between $\textbf{E}_{e \leftarrow e2c}^{s}$ and $\textbf{E}_{e \leftarrow e2t}^{d}$\footnote{The same process  can be applied between $\textbf{E}_{t \leftarrow tct}^{s}$ and $\textbf{E}_{t \leftarrow e2t}^{d}$, and  $\textbf{E}_{c \leftarrow tct}^{s}$ and $\textbf{E}_{c \leftarrow e2c}^{d}$}. To ease  notation, we respectively use $\textbf{E}^{s}$ and $\textbf{E}^{d}$ instead of $\textbf{E}_{e \leftarrow e2c}^{s}$ and  $\textbf{E}_{e \leftarrow e2t}^{d}$. 
The process to align different embeddings follows three steps:}
\begin{enumerate}[itemsep=0.5ex, leftmargin=4mm]
\item \textit{Construction of  Probability Distributions.}
\vic{We first construct the distributions $\bm{\mu}$ and $\bm{\nu}$ for the entity embeddings $\textbf{E}^{s}$ and  $\textbf{E}^{d}$, respectively. We define  $\bm{\mu}=\sum_{i=1}^{n}\alpha_i\delta(\textbf{E}^{s}_i)$ and $\bm{\nu}=\sum_{j=1}^{m}\beta_j\delta(\textbf{E}^{d}_j)$, where  $\textbf{E}_i^s$ and $\textbf{E}_j^d$  respectively represent the values at the $i$-th and $j$-th positions in the embeddings $\textbf{E}^{s}$ and $\textbf{E}^{d}$. Further, the weight vectors $\alpha_i$  and $\beta_j$ satisfy $\sum_{i=1}^n \alpha_i=\sum_{j=1}^m \beta_j=1$, and to simplify the calculations, we set $\alpha_i=1/n$ and  $\beta_j=1/m$. 
}


\item \textit{Transport Scheme.} 
\vic{Based on  the distributions $\bm{\mu}$ and $\bm{\nu}$,  we can obtain all joint probability distributions $\Pi(\bm{\mu}, \bm{\nu})$. At the same time, using the cosine distance, we can calculate the distance $\textbf{C}_{ij}=1-\frac{{\textbf{E}_i^s}^{\top}\textbf{E}_j^d}{\|\textbf{E}_i^s\|_2\|\textbf{E}_j^d\|_2}$ between the elements of $\textbf{E}_i^s$ and $\textbf{E}_j^d$. Further, if $\Pi(\bm{\mu}, \bm{\nu})$ and $\textbf{C}_{ij}$ are known, we can apply the Sinkhorn algorithm~\cite{Marco_2013} to optimize Eq.\ \eqref{equation_1} and  obtain the optimal transportation matrix \textbf{T}.}


\item \textit{Translate Embedding Representation.}
\vic{We multiply the source entity embedding $\textbf{E}^{s}$ with the transportation matrix \textbf{T} to get the aligned entity embedding $\widehat{\textbf{E}^{s}}={\textbf{E}^{s}}^{\top}\textbf{T}$. Further, we add $\widehat{\textbf{E}^{s}}$ with the destination entity embedding $\textbf{E}^{d}$ to get the final entity embedding representation $\textbf{Z}^{e}$. Following the same steps, we can obtain the cross-view aligned type embeddings $\textbf{Z}^{t}$ and cluster embeddings $\textbf{Z}^{c}$.}

\end{enumerate}
\subsection{Pooling-based Entity Typing Prediction}
\smallskip \noindent \textbf{Single Neighbor Prediction.}
\vic{After introducing cluster information into the KGET task, the neighbors of an entity  include relational, type and cluster neighbors. We will collectively refer to them as entity neighbors, and  will concatenate their embedding representations to obtain a unified embedding $\textbf{Z}=[\textbf{Z}^{e},\textbf{Z}^{t},\textbf{Z}^{c}]$. The neighbors of an entity provide supportive evidence for predicting entity types, wherein different neighbors contribute differently to the inference of the types of an entity. We introduce the \emph{single neighbor prediction} mechanism, which  focuses on a single neighbor during inference, reducing the interference from irrelevant information from other neighbors. For the $i$-th neighbor $f_i$ of an entity $e$, we define $\mathcal{N}_i=\{(r, f_i)| (e, r, f_i) \in \mathcal{K}\}$\footnote{We assume a fixed, but arbitrary linear order on $\mathcal N_e^\mathcal G \cup \mathcal N_e^\mathcal T$.}.
We then follow the translation method TransE~\cite{Antoine_2013} and apply a \emph{multilayer perceptron} to obtain the neighbor embedding $\bm{\mathcal{N}}_i=\textbf{W}Relu(\textbf{f}_i-\textbf{r})+\textbf{b}$, where $\textbf{f}_i$ and $\textbf{r}$ are the embedding representations of $f_i$ and $r$, respectively, $\textbf{W} \in \mathbb{R}^{N\times D}$, $\textbf{b} \in \mathbb{R}^N$ are the learnable parameters, $D$ represents the dimension of the embedding, and $N$ represents the number of types in the input KG. }

\smallskip \noindent \textbf{Mixture Pooling Strategy.} \vic{For an entity $e$ with $L$ neighbors, we  get the final predicted relevance score for all neighbors: $\bm{\mathcal{N}}_e=\{\bm{\mathcal{N}}_1,\bm{\mathcal{N}}_2, ..., \bm{\mathcal{N}}_L\}$. To aggregate multiple entity typing inference results from different neighbors, we introduce a \emph{mixture pooling module (MPM)}, containing three pooling sub-modules, namely, max pooling $\textbf{S}^m=max(\bm{\mathcal{N}}_e)$, average pooling $\textbf{S}^a=avg(\bm{\mathcal{N}}_e)$, and multi-head weight pooling $\textbf{S}^w$, defined as follows:
\begin{equation}
\label{equation_2}
\textbf{S}^w=\sum_{i=1}^{H}\sum_{j=1}^{L}w_i\bm{\mathcal{N}}_j,\,\,
w_i=\frac{{\rm exp}\,(h_i\bm{\mathcal{N}}_j)}
{\sum_{k=1}^{L}{\rm exp}\,(h_i\bm{\mathcal{N}}_k)}
\end{equation}
where \textit{H} represents the number of heads, and $h_i$ represents the temperature value of the $i$-th head, which is a  scalar hyperparameter. We merge the three pooling results through addition, and use the sigmoid function $\sigma$ to project the value to the range 0 to 1. We finally define $\textbf{S}_e=\sigma(\textbf{S}^w+\textbf{S}^m+\textbf{S}^a)$, where the $p$-th element $\textbf{s}_{e, p}$ represents the probability that the entity $e$ has type $p$.}


\begin{table*}[!htp]
\renewcommand\arraystretch{1.1}
\setlength{\tabcolsep}{1.20em}
\centering
\small
\begin{tabular*}{0.69\linewidth}{@{}ccccccc@{}}
\hline
\multicolumn{1}{c|}{\textbf{Datasets}} & \multicolumn{3}{c|}{\textbf{FB15kET}} & \multicolumn{3}{c}{\textbf{YAGO43kET}}\\
\hline
\multicolumn{1}{c|}{\textbf{Statistics}} & \multicolumn{1}{c}{\textbf{full}} & \multicolumn{1}{c}{\textbf{easy}} & \multicolumn{1}{c|}{\textbf{hard}} & \multicolumn{1}{c}{\textbf{full}} & \multicolumn{1}{c}{\textbf{easy}} & \multicolumn{1}{c}{\textbf{hard}} \\
\hline
\multicolumn{1}{l|}{\# Entities}   &14,951  &14,951 &\multicolumn{1}{c|}{14,951}  &42,335 &42,335  &42,335\\
\multicolumn{1}{l|}{\# Relations}   &1,345  &1,345  &\multicolumn{1}{c|}{1,345}  &37 &37  &37 \\
\multicolumn{1}{l|}{\# Types}   &3,584  &676  &\multicolumn{1}{c|}{2,908}  &45,182 &3,925  &41,257 \\
\multicolumn{1}{l|}{\# Clusters}   &1,081  &1,081    &\multicolumn{1}{c|}{1,081}  &1,124  &1,124  &1,124 \\
\multicolumn{1}{l|}{\# Train.triples}   &483,142  &483,142 &\multicolumn{1}{c|}{483,142}  &331,686  &331,686  &331,686 \\
\multicolumn{1}{l|}{\# Train.tuples}   &136,618  &126,950    &\multicolumn{1}{c|}{9,668}  &375,853  &263,276  &112,577 \\
\multicolumn{1}{l|}{\# Valid.tuples}   &15,848  &14,736  &\multicolumn{1}{c|}{1,112}  &43,111    &31,532  &11,579 \\
\multicolumn{1}{l|}{\# Test.tuples}   &15,847  &14,825   &\multicolumn{1}{c|}{1,022}  &43,119    &31,672  &11,447 \\
\hline
\end{tabular*}
\caption{Statistics of Datasets. \#Train.triples represents the number of edges of relational graph $\mathcal{G}$, \#Train.tuples + \#Valid.tuples + \#Test.tuples represents the number of edges of type graph $\mathcal{T}$.}
\label{table_datasets}
\end{table*}

\smallskip \noindent \textbf{Prediction and Optimization.} 
\vic{In principle, one could choose the well-known binary cross-entropy loss  for model training, by treating all pairs ($e, has\_type, t$) occurring in $\mathcal{T}$ as positive examples and considering all pairs ($e, has\_type, t$) not occurring in $\mathcal{T}$ as negative samples. However, due to the inherent incompleteness of the input knowledge graph, many false negative samples are introduced since certain  pairs ($e, has\_type, t$) might be absent due to omissions rather than their invalidity. Certainly, the missing types of entities that we aim at inferring fall into this category. To mitigate the issue of false negatives, we propose a \emph{Beta distribution-based  cross-entropy (BDCE)} loss (cf.\ Equation~\eqref{equation_4}),  based on the Beta probability distribution.}
\begin{eqnarray}
\label{equation_4}
\mathcal{L}_{\text{BDCE}}&=&-\sum_{(e,p)\notin\mathcal{T}}{\theta \cdot f_{(\alpha, \beta)}(\textbf{s}_{e,p}')\rm log}(1-\textbf{s}_{e,p}') \nonumber     \\
&\;&-\sum_{(e,p)\in\mathcal{T}}{\rm log}(\textbf{s}_{e,p}) \nonumber \\ \vspace{-0.5mm}
\end{eqnarray}
\vic{where $\theta$ is a hyperparameter used to regulate the overall weight of negative samples, $\textbf{s}_{e,p}$ and $\textbf{s}_{e,p}'$ represent the positive score and negative score, respectively. The Beta distribution has two types of hyperparameters $\alpha$ and $\beta$, its probability density function (PDF) is defined as: $f_{(\alpha, \beta)}(x)=\frac{x^{\alpha-1}(1-x)^{\beta-1}}{\textbf{B}(\alpha, \beta)}$, where $x \in [0,1]$ and $\textbf{B}(\cdot)$ denotes the Beta function. Negative samples with higher scores are more likely to be false negatives, while negative samples with lower scores are considered as easier samples to learn. $f_{(\alpha, \beta)}$ will assign lower weight coefficient to those negative samples with excessively high or low scores, which can alleviate the false negatives problem and guide the model to learn more challenging hard samples.}

\section{Experiments}
\vic{To evaluate the effectiveness of our model, we aim to explore the following four research questions:}
\begin{itemize}[itemsep=0.5ex, leftmargin=5mm]
\item
\vic{\textbf{RQ1 (Effectiveness):} How our model performs compared to the state-of-the-art under different conditions?}
\item 
\vic{\textbf{RQ2 (Ablation studies):} How different components  contribute to COTET's performance?}
\item 
\vic{\textbf{RQ3 (Parameter sensitivity):} How hyper-parameters influence COTET's performance?}
\item
\vic{\textbf{RQ4 (Complexity analysis):} What is the amount of computation and parameters used by COTET?}
\end{itemize}

\begin{table}[!hp]
\renewcommand\arraystretch{1.1}
\setlength{\tabcolsep}{0.30em}
\centering
\small
\begin{tabular*}{0.9\linewidth}{@{}cc@{}}
\toprule
\multicolumn{1}{l}{\textbf{Parameter}} & \multicolumn{1}{c}{\textbf{\{FB15kET,\,\, YAGO43kET\}}} \\
\midrule
\multicolumn{1}{l}{\# Embedding dimensions}   &\multicolumn{1}{c}{\{100,\,\,\,100\}} \\
\multicolumn{1}{l}{\# Learning rate}   &\multicolumn{1}{c}{\{0.001,\,\,\,0.001\}} \\
\multicolumn{1}{l}{\# LightGCN layers}   &\multicolumn{1}{c}{\{4,\,\,\,1\}} \\
\multicolumn{1}{l}{\# CompGCN layers}   &\multicolumn{1}{c}{\{2,\,\,\,2\}} \\
\multicolumn{1}{l}{\# Number of heads}   &\multicolumn{1}{c}{\{5,\,\,\,5\}} \\
\multicolumn{1}{l}{\# $\alpha$ value}   &\multicolumn{1}{c}{\{2.0,\,\,\,2.0\}} \\
\multicolumn{1}{l}{\# $\beta$ value}   &\multicolumn{1}{c}{\{2.0,\,\,\,2.0\}} \\
\multicolumn{1}{l}{\# $\theta$ value}   &\multicolumn{1}{c}{\{0.7,\,\,\,0.5\}} \\
\bottomrule
\end{tabular*}
\caption{The best hyperparameters settings of the FB15kET and YAGO43kET datasets.  }
\label{table_hyperparameters}
\end{table}

\subsection{Experimental Setup}
\smallskip \noindent \textbf{Datasets.} 
\vic{We conduct experiments on the following three different versions of the FB15kET and YAGO43kET datasets to verify the effectiveness and robustness of COTET:}
\begin{itemize}[itemsep=0.5ex, leftmargin=5mm]
\item
\vic{\textit{Full Version}: We evaluate COTET on two well-known KGs, each composed of a relational  graph $\mathcal{G}$ and type graph $\mathcal{T}$. For $\mathcal{G}$, we select the KGs like FB15k~\cite{Antoine_2013} and YAGO43k~\cite{Moon_2017}. For $\mathcal{T}$, we use FB15kET and YAGO43kET~\cite{Pan_2021}, which map entities from FB15k and YAGO43k to their corresponding entity types. The statistical results of the datasets are shown in Table~\ref{table_datasets}.}
\item 
\vic{\textit{Hard and Easy Versions}: We constructed \textit{hard} and \textit{easy} datasets based on how frequently entity types appear in the datasets. Given a threshold $k$, a type will be considered hard if occurs $m \leq k$ times in a dataset, while it is considered easy if it occurs $n > k$ times. A dataset is easy/hard if all its types are easy/hard. The frequency thresholds $k$ of FB15kET and YAGO43kET are set to 15 and 5, respectively. The statistical results of the datasets are shown in Table~\ref{table_datasets}.}
\item 
\vic{\textit{Sparse Neighbor-connection Versions}: In real-world KGs, many entities have sparse connections to other entities, which may result in the lack of  structural knowledge. Especially for the KGET task, we need to use the content of neighbor entities to predict the possible types of an entity. 
We look at the impact of sparse entity neighbors on the entity type prediction to explore the robustness of COTET under different extreme conditions. As an instance,  take  the FB15kET dataset, we construct two versions of neighbor-sparse sub-datasets: \textit{different dropping rates of relational neighbors} and \textit{different dropping rates of relational types}, which randomly remove 25\%, 50\%, 75\%, and 90\% of the  relational neighbors and relational types, respectively,  in the relational graph $\mathcal{G}$  of a given  entity. Note that  to understand the impact of sparsity, we only modify the relational graph $\mathcal{G}$, keeping the type graph $\mathcal{T}$ unchanged.} \vic{To illustrate the difference between the two neighbor-sparse subsets, consider the following example. For the entity \textit{Lionel\_Messi}, there are four triples (\textit{Lionel\_Messi}, \textit{member\_of\_sports\_team}, \textit{FC\_Barcelona}), (\textit{Lionel\_Messi}, \textit{member\_of\_sports\_team}, \textit{Paris\_Saint-Germain}), (\textit{Lionel\_Messi}, \textit{teammate}, \textit{Neymar}), and (\textit{Lionel\_Messi}, \textit{teammate}, \textit{Kylian Mbappé}). Dropping relational neighbors will delete any of the four (e.g.\ (\textit{Lionel\_Messi}, \textit{member\_of\_sports\_team}, \textit{FC\_Barcelona}) and (\textit{Lionel\_Messi}, \textit{teammate}, \textit{Kylian Mbappé})), while dropping relational types will delete e.g.\ all triples under relational type \textit{member\_of\_sports\_team}: (\textit{Lionel\_Messi}, \textit{member\_of\_sports\_team}, \textit{FC\_Barcelona}) and (\textit{Lionel\_Messi}, \textit{member\_of\_sports\_team}, \textit{Paris\_Saint-Germain})).} 
\end{itemize}


\smallskip \noindent \textbf{Implementation Details.} We use Adam~\cite{Diederik_2014} optimizer to update the parameters. The hyperparameters are validated in the following range, embedding dimensions from $\{50, 100, 150\}$, the number of LightGCN and CompGCN layers from $\{1, 2, 3, 4\}$, the learning rate from $\{0.001, 0.005, 0.01\}$, the number of heads $H \in \{3, 4, 5\}$, the weight of negative samples in BDCE loss $\theta \in \{0.5, 0.6, 0.7, 0.8\}$, $\alpha$ and $\beta$ parameters in the Beta function from $\{1.5, 2.0, 2.5, 3.0\}$. Table~\ref{table_hyperparameters} summarizes the best hyperparameters settings in FB15kET and YAGO43kET datasets.

\smallskip \noindent \textbf{Evaluation Protocol.} We evaluate the  performance on the KGET task using two ranking-based metrics: \textit{i}) \emph{mean reciprocal rank (MRR)}, which measures the mean of inverse ranks assigned to correct types; \textit{ii}) \emph{Hits@$K$} ($K \in\{1, 3, 10\}$), which measures the proportion of correct types among the top $K$ predicted types. All metrics follow the larger the value, the better the effect. Following the state-of-the-art baselines~\cite{Pan_2021, Hu_2022, Jin_2022}, we also adopt the filtered setting~\cite{Antoine_2013} to remove all the known types during evaluation.

\begin{table*}[!htp]
\renewcommand\arraystretch{1.15}
\setlength{\tabcolsep}{0.75em}
\centering
\small
\begin{tabular*}{0.83\linewidth}{@{}cccccccccc@{}}
\hline 
\multicolumn{1}{c|}{\textbf{Datasets}} & \multicolumn{4}{c|}{\textbf{FB15kET}} & \multicolumn{4}{c}{\textbf{YAGO43kET}}\\ 
\hline
\multicolumn{1}{c|}{\textbf{Metrics}} & \multicolumn{1}{c}{\textbf{MRR}} & \multicolumn{1}{c}{\textbf{Hits@1}} & \multicolumn{1}{c}{\textbf{Hits@3}} & \multicolumn{1}{c|}{\textbf{Hits@10}} & \multicolumn{1}{c}{\textbf{MRR}} & \multicolumn{1}{c}{\textbf{Hits@1}} & \multicolumn{1}{c}{\textbf{Hits@3}} & \multicolumn{1}{c}{\textbf{Hits@10}} \\
\hline
\multicolumn{9}{c}{\textit{Embedding-based methods}} \\
\hline
\multicolumn{1}{l|}{TransE~\cite{Antoine_2013}$^\blacklozenge$} &0.618  &0.504  &0.686  &\multicolumn{1}{c|}{0.835}  &0.427  &0.304  &0.497  &0.663 \\
\multicolumn{1}{l|}{ComplEx~\cite{Trouillon_2016}$^\blacklozenge$} &0.595  &0.463  &0.680  &\multicolumn{1}{c|}{0.841}  &0.435  &0.316  &0.504  &0.658 \\
\multicolumn{1}{l|}{RotatE~\cite{Zhiqing_2019}$^\blacklozenge$} &0.632  &0.523  &0.699  &\multicolumn{1}{c|}{0.840}  &0.462  &0.339  &0.537  &0.695 \\
\multicolumn{1}{l|}{CompoundE~\cite{Xiou_2023}$^\blacklozenge$} &0.640  &0.525  &0.719  &\multicolumn{1}{c|}{0.859}  &0.480  &0.364  &0.558  &0.703 \\
\multicolumn{1}{l|}{SimKGC~\cite{Liang_2022}$^\blacklozenge$} &0.317  &0.210  &0.348  &\multicolumn{1}{c|}{0.545}  &0.172  &0.097  &0.184  &0.317 \\
\multicolumn{1}{l|}{ETE~\cite{Moon_2017}$^\lozenge$} &0.500  &0.385  &0.553  &\multicolumn{1}{c|}{0.719}  &0.230  &0.137  &0.263  &0.422 \\
\multicolumn{1}{l|}{ConnectE~\cite{Zhao_2020}$^\lozenge$} &0.590  &0.496  &0.643  &\multicolumn{1}{c|}{0.799}  &0.280  &0.160  &0.309  &0.479 \\
\multicolumn{1}{l|}{CORE~\cite{Ge_2021}$^\lozenge$} &0.600  &0.489  &0.663  &\multicolumn{1}{c|}{0.816}  &0.350  &0.242  &0.392  &0.550 \\
\hline
\multicolumn{9}{c}{\textit{GNN-based methods}} \\
\hline
\multicolumn{1}{l|}{HMGCN~\cite{Jin_2019}$^\blacklozenge$} &0.510  &0.390  &0.548  &\multicolumn{1}{c|}{0.724}  &0.250  &0.142  &0.273  &0.437 \\
\multicolumn{1}{l|}{AttEt~\cite{Zhuo_2022}$^\lozenge$} &0.620  &0.517  &0.677  &\multicolumn{1}{c|}{0.821}  &0.350  &0.244  &0.413  &0.565 \\
\multicolumn{1}{l|}{ConnectE-MRGAT~\cite{Zhao_2022}$^\lozenge$} &0.630  &0.562  &0.662  &\multicolumn{1}{c|}{0.804}  &0.320  &0.243  &0.343  &0.482 \\
\multicolumn{1}{l|}{RACE2T~\cite{Zou_2022}$^\lozenge$} &0.640  &0.561  &0.689  &\multicolumn{1}{c|}{0.817}  &0.340  &0.248  &0.376  &0.523 \\
\multicolumn{1}{l|}{CompGCN~\cite{Vashishth_2020}$^\blacklozenge$} &0.665  &0.578  &0.712  &\multicolumn{1}{c|}{0.839}  &0.355  &0.274  &0.383  &0.513 \\
\multicolumn{1}{l|}{RGCN~\cite{Michael_2018}$^\lozenge$} &0.679  &0.597  &0.722  &\multicolumn{1}{c|}{0.843}  &0.372  &0.281  &0.409  &0.549 \\
\multicolumn{1}{l|}{CET~\cite{Pan_2021}$^\lozenge$} &0.697  &0.613  &0.745  &\multicolumn{1}{c|}{0.856}  &0.503  &0.398  &0.567  &0.696 \\
\multicolumn{1}{l|}{MiNer~\cite{Jin_2022}$^\lozenge$} &0.728  &0.654  &0.768  &\multicolumn{1}{c|}{0.875}  &0.521  &0.412  &0.589  &0.714 \\
\hline
\multicolumn{9}{c}{\textit{Transformer-based methods}} \\
\hline
\multicolumn{1}{l|}{CoKE~\cite{Wang_2019}$^\blacklozenge$} &0.465  &0.379  &0.510  &\multicolumn{1}{c|}{0.624}  &0.344  &0.244  &0.387  &0.542 \\
\multicolumn{1}{l|}{HittER~\cite{Chen_2021}$^\blacklozenge$} &0.422  &0.333  &0.466  &\multicolumn{1}{c|}{0.588}  &0.240  &0.163  &0.259  &0.390 \\
\multicolumn{1}{l|}{TET~\cite{Hu_2022}$^\lozenge$} &0.717  &0.638  &0.762  &\multicolumn{1}{c|}{0.872}  &0.510  &0.408  &0.571  &0.695 \\
\hline
\multicolumn{9}{c}{\textit{Our methods}} \\
\hline
\multicolumn{1}{c|}{$\rm COTET_1$} &0.754  &0.683  &0.793  &\multicolumn{1}{c|}{0.893}  &0.494  &0.388  &0.559  &0.689 \\
\multicolumn{1}{c|}{$\rm COTET_2$} &\textbf{0.762}  &\textbf{0.694}  &\textbf{0.801}  &\multicolumn{1}{c|}{\textbf{0.895}}  &\textbf{0.529}  &\textbf{0.419}  &\textbf{0.597}  &\textbf{0.729} \\
\hline
\end{tabular*}
\caption{Results for the KGET task on FB15kET and YAGO43kET datasets. [$\lozenge$]: Results are taken from the original papers. [$\blacklozenge$] results are from our implementation of the corresponding models. Best scores are highlighted in \textbf{bold}. $\rm COTET_2$ denotes the normal version  described in \S\ref{sec:method}, and $\rm COTET_1$ denotes the version obtained after reversing the source and destination embeddings. Unless otherwise specified, COTET means $\rm COTET_2$.}
\label{table_main_result}
\end{table*}

\smallskip \noindent \vic{\textbf{Baselines.} We compare COTET with three kinds of baselines for KGET, including \textit{embedding-based models}, \textit{GNN-based models} and \textit{transformer-based models}. For \textit{embedding-based models}, we  consider the following methods: TransE~\cite{Antoine_2013}, ComplEx~\cite{Trouillon_2016}, RotatE~\cite{Zhiqing_2019}, CompoundE~\cite{Xiou_2023}, SimKGC~\cite{Liang_2022}, ETE~\cite{Moon_2017}, ConnectE~\cite{Zhao_2020}, and CORE~\cite{Ge_2021}. For \textit{GNN-based models}, we select  HMGCN~\cite{Jin_2019}, AttEt~\cite{Zhuo_2022}, ConnectE-MRGAT~\cite{Zhao_2022}, RACE2T~\cite{Zou_2022}, CompGCN~\cite{Vashishth_2020}, RGCN~\cite{Pan_2021}, CET~\cite{Pan_2021}, and MiNer~\cite{Jin_2022}. For \textit{transformer-based models}, we consider  CoKE~\cite{Wang_2019}, HittER~\cite{Chen_2021}, and TET~\cite{Hu_2022}.}




\subsection{Main Results}
\vic{To address \textbf{RQ1}, we conduct experiments on three variants of the  the FB15kET and YAGO43kET datasets. \textit{(i)}: full version, in which experiments are conducted using the full datasets FB15kET and YAGO43kET. The corresponding results are shown in Table~\ref{table_main_result}; \textit{(ii)} hard and easy versions of FB15kET and YAGO43kET, the corresponding performance  results are presented in Table~\ref{table_easy_hard_subsets}. \textit{(iii)} sparse neighbor-connections  versions  of FB15kET and YAGO43kET, the corresponding performance results are reported in Table~\ref{table_ablation_with_different_relational_neighbors_dropping_rates} and Table~\ref{table_ablation_with_different_relation_type_dropping_rates}.}

\smallskip \noindent \textbf{Full Version.} From the results of Table~\ref{table_main_result}, we can draw the following conclusions. On the one hand, COTET achieves the best performance on the  KGET task, significantly outperforming (in all evaluation metrics) existing SoTA baselines. $\rm COTET_2$  respectively brings a 3.4\% and 0.8\%  improvement on MRR in the FB15kET and YAGO43kET datasets over MiNer (the best baseline). These results confirm the importance of the interaction between the proposed modules. On the other hand, we observe that swapping source and destination embeddings in the cross-view optimal transport module affects the magnitude of 
improvement. The performance of the corresponding model $\rm COTET_1$ is substantially lower than that of the normal version $\rm COTET_2$: we observe that $\rm COTET_1$ respectively drops 0.8\% and 3.5\% on the MRR metric over the FB15kET and YAGO43kET datasets. This can be intuitively explained by the fact that in the optimal transport mechanism, the role of the source  and destination distributions are different, and in practice they cannot be swapped. One needs to carefully select which the source and  destination roles according to the characteristics of the task.

\smallskip \noindent \textbf{Hard and Easy Versions.}
\vic{We generate easy and hard versions of the considered datasets as explained above. Since the types in the hard version appear less frequently, they are more difficult to predict. Remarkably, we observe in  Table~\ref{table_easy_hard_subsets} that  in both FB15kET and YAGO43kET,  COTET  shows a higher improvement in the hard version than in the easy one. Specifically, compared to  MiNer, COTET respectively achieves an improvement of 3.0\% and 6.0\% in terms of the MRR metric on the FB15kET-easy and the FB15kET-hard datasets. This improvement is mainly attributed to the introduction of cluster information, which provides additional semantic knowledge to the original entity-type view,  providing a stronger guidance for types with fewer samples in the original view.}

\begin{table*}
\renewcommand\arraystretch{1.15}
\setlength{\tabcolsep}{0.565em}
\centering
\small
\begin{tabular*}{0.93\linewidth}{@{}ccccccccccccc@{}}
\hline
\multicolumn{1}{c|}{} & \multicolumn{3}{c|}{\textbf{FB15kET-easy}} & \multicolumn{3}{c|}{\textbf{FB15kET-hard}} & \multicolumn{3}{c|}{\textbf{YAGO43kET-easy}} &\multicolumn{3}{c}{\textbf{YAGO43kET-hard}}\\
\hline
\multicolumn{1}{c|}{\textbf{Models}}  & \multicolumn{1}{c}{\textbf{MRR}} & \multicolumn{1}{c}{\textbf{Hits@1}} & \multicolumn{1}{c|}{\textbf{Hits@3}} & \multicolumn{1}{c}{\textbf{MRR}} & \multicolumn{1}{c}{\textbf{Hits@1}} & \multicolumn{1}{c|}{\textbf{Hits@3}} & \multicolumn{1}{c}{\textbf{MRR}} & \multicolumn{1}{c}{\textbf{Hits@1}} & \multicolumn{1}{c|}{\textbf{Hits@3}} & \multicolumn{1}{c}{\textbf{MRR}} & \multicolumn{1}{c}{\textbf{Hits@1}} & \multicolumn{1}{c}{\textbf{Hits@3}} \\
\hline
\multicolumn{1}{c|}{CompGCN} 
&0.699  &0.612  &\multicolumn{1}{c|}{0.747}
&0.441  &0.344  &\multicolumn{1}{c|}{0.495}
&0.452  &0.352  &\multicolumn{1}{c|}{0.492}
&0.217  &0.148  &\multicolumn{1}{c}{0.238} \\
\multicolumn{1}{c|}{RGCN} 
&0.711  &0.627  &\multicolumn{1}{c|}{0.754}
&0.435  &0.345  &\multicolumn{1}{c|}{0.480}
&0.463  &0.355  &\multicolumn{1}{c|}{0.512}
&0.219  &0.154  &\multicolumn{1}{c}{0.236} \\
\multicolumn{1}{c|}{CET} 
&0.731  &0.646  &\multicolumn{1}{c|}{0.782}
&0.480  &0.387  &\multicolumn{1}{c|}{0.526}
&0.626  &0.510  &\multicolumn{1}{c|}{0.705}
&0.273  &0.206  &\multicolumn{1}{c}{0.295} \\
\multicolumn{1}{c|}{TET} 
&0.751  &0.674  &\multicolumn{1}{c|}{0.795}
&0.390  &0.316  &\multicolumn{1}{c|}{0.427}
&0.622  &0.513  &\multicolumn{1}{c|}{0.696}
&0.287  &0.229  &\multicolumn{1}{c}{0.311} \\
\multicolumn{1}{c|}{MiNer} 
&0.758  &0.684  &\multicolumn{1}{c|}{0.801}
&0.512  &0.426  &\multicolumn{1}{c|}{0.551}
&0.638  &0.520  &\multicolumn{1}{c|}{0.722}
&0.299  &0.220  &\multicolumn{1}{c}{0.328} \\
\hline
\multicolumn{1}{c|}{COTET} 
&\textbf{0.788}  &\textbf{0.721}  &\multicolumn{1}{c|}{\textbf{0.826}}
&\textbf{0.572}  &\textbf{0.483}  &\multicolumn{1}{c|}{\textbf{0.614}}
&\textbf{0.649}  &\textbf{0.529}  &\multicolumn{1}{c|}{\textbf{0.735}}
&\textbf{0.317}  &\textbf{0.239}  &\multicolumn{1}{c}{\textbf{0.348}} \\
\hline
\end{tabular*}
\caption{Results  on easy and hard variants of the FB15kET and YAGO43kET datatsets. Best scores are highlighted in \textbf{bold}.}
\label{table_easy_hard_subsets}
\end{table*}

\begin{table*}[!htp]
\renewcommand\arraystretch{1.15}
\setlength{\tabcolsep}{0.28em}
\centering
\small
\begin{tabular*}{0.855\linewidth}{@{}ccccccccccccc@{}}
\hline
\multicolumn{1}{c|}{\textbf{Dropping Rates}} & \multicolumn{3}{c|}{\textbf{25\%}} & \multicolumn{3}{c|}{\textbf{50\%}} & \multicolumn{3}{c|}{\textbf{75\%}} &\multicolumn{3}{c}{\textbf{90\%}}\\
\hline
\multicolumn{1}{c|}{\textbf{Models}}  & \multicolumn{1}{c}{\textbf{MRR}} & \multicolumn{1}{c}{\textbf{Hits@1}} & \multicolumn{1}{c|}{\textbf{Hits@3}} & \multicolumn{1}{c}{\textbf{MRR}} & \multicolumn{1}{c}{\textbf{Hits@1}} & \multicolumn{1}{c|}{\textbf{Hits@3}} & \multicolumn{1}{c}{\textbf{MRR}} & \multicolumn{1}{c}{\textbf{Hits@1}} & \multicolumn{1}{c|}{\textbf{Hits@3}} & \multicolumn{1}{c}{\textbf{MRR}} & \multicolumn{1}{c}{\textbf{Hits@1}} & \multicolumn{1}{c}{\textbf{Hits@3}}\\
\hline
\multicolumn{1}{c|}{CompGCN~\cite{Vashishth_2020}} 
&0.661  &0.573  &\multicolumn{1}{c|}{0.705}
&0.655  &0.565  &\multicolumn{1}{c|}{0.702}
&0.648  &0.559  &\multicolumn{1}{c|}{0.697}
&0.633  &0.544  &\multicolumn{1}{c}{0.679} \\
\multicolumn{1}{c|}{RGCN~\cite{Michael_2018}} 
&0.673  &0.590  &\multicolumn{1}{c|}{0.716}
&0.667  &0.584  &\multicolumn{1}{c|}{0.708}
&0.648  &0.560  &\multicolumn{1}{c|}{0.694}
&0.626  &0.534  &\multicolumn{1}{c}{0.673} \\
\multicolumn{1}{c|}{CET~\cite{Pan_2021}} 
&0.697  &0.613  &\multicolumn{1}{c|}{0.744}
&0.687  &0.601  &\multicolumn{1}{c|}{0.733}
&0.670  &0.580  &\multicolumn{1}{c|}{0.721}
&0.646  &0.553  &\multicolumn{1}{c}{0.698} \\
\multicolumn{1}{c|}{TET~\cite{Hu_2022}} 
&0.712  &0.631  &\multicolumn{1}{c|}{0.758}
&0.705  &0.624  &\multicolumn{1}{c|}{0.753}
&0.689  &0.606  &\multicolumn{1}{c|}{0.733}
&0.658  &0.574  &\multicolumn{1}{c}{0.701} \\
\multicolumn{1}{c|}{MiNer~\cite{Jin_2022}} 
&0.714  &0.634  &\multicolumn{1}{c|}{0.760}
&0.703  &0.620  &\multicolumn{1}{c|}{0.749}
&0.683  &0.596  &\multicolumn{1}{c|}{0.731}
&0.652  &0.556  &\multicolumn{1}{c}{0.706} \\
\hline
\multicolumn{1}{c|}{COTET} 
&\textbf{0.752}  &\textbf{0.682}  &\multicolumn{1}{c|}{\textbf{0.788}}
&\textbf{0.746}  &\textbf{0.674}  &\multicolumn{1}{c|}{\textbf{0.786}}
&\textbf{0.726}  &\textbf{0.649}  &\multicolumn{1}{c|}{\textbf{0.768}}
&\textbf{0.705}  &\textbf{0.624}  &\multicolumn{1}{c}{\textbf{0.748}} \\
\hline
\end{tabular*}
\caption{Evaluation with different relational neighbors dropping rates on FB15kET. Best scores are highlighted in \textbf{bold}.}
\label{table_ablation_with_different_relational_neighbors_dropping_rates}
\end{table*}

\begin{table*}[!htp]
\renewcommand\arraystretch{1.15}
\setlength{\tabcolsep}{0.28em}
\centering
\small
\begin{tabular*}{0.855\linewidth}{@{}ccccccccccccc@{}}
\hline
\multicolumn{1}{c|}{\textbf{Dropping Rates}} & \multicolumn{3}{c|}{\textbf{25\%}} & \multicolumn{3}{c|}{\textbf{50\%}} & \multicolumn{3}{c|}{\textbf{75\%}} &\multicolumn{3}{c}{\textbf{90\%}}\\
\hline
\multicolumn{1}{c|}{\textbf{Models}}  & \multicolumn{1}{c}{\textbf{MRR}} & \multicolumn{1}{c}{\textbf{Hits@1}} & \multicolumn{1}{c|}{\textbf{Hits@3}} & \multicolumn{1}{c}{\textbf{MRR}} & \multicolumn{1}{c}{\textbf{Hits@1}} & \multicolumn{1}{c|}{\textbf{Hits@3}} & \multicolumn{1}{c}{\textbf{MRR}} & \multicolumn{1}{c}{\textbf{Hits@1}} & \multicolumn{1}{c|}{\textbf{Hits@3}} & \multicolumn{1}{c}{\textbf{MRR}} & \multicolumn{1}{c}{\textbf{Hits@1}} & \multicolumn{1}{c}{\textbf{Hits@3}}\\
\hline
\multicolumn{1}{c|}{CompGCN~\cite{Vashishth_2020}} 
&0.664  &0.578  &\multicolumn{1}{c|}{0.708}
&0.662  &0.574  &\multicolumn{1}{c|}{0.708}
&0.653  &0.565  &\multicolumn{1}{c|}{0.699}
&0.637  &0.546  &\multicolumn{1}{c}{0.683} \\
\multicolumn{1}{c|}{RGCN~\cite{Michael_2018}} 
&0.676  &0.593  &\multicolumn{1}{c|}{0.719}
&0.673  &0.590  &\multicolumn{1}{c|}{0.719}
&0.658  &0.573  &\multicolumn{1}{c|}{0.702}
&0.636  &0.548  &\multicolumn{1}{c}{0.681} \\
\multicolumn{1}{c|}{CET~\cite{Pan_2021}} 
&0.699  &0.617  &\multicolumn{1}{c|}{0.743}
&0.694  &0.610  &\multicolumn{1}{c|}{0.742}
&0.675  &0.588  &\multicolumn{1}{c|}{0.721}
&0.653  &0.564  &\multicolumn{1}{c}{0.700} \\
\multicolumn{1}{c|}{TET~\cite{Hu_2022}} 
&0.711  &0.631  &\multicolumn{1}{c|}{0.756}
&0.710  &0.630  &\multicolumn{1}{c|}{0.757}
&0.690  &0.608  &\multicolumn{1}{c|}{0.734}
&0.677  &0.591  &\multicolumn{1}{c}{0.722} \\
\multicolumn{1}{c|}{MiNer~\cite{Jin_2022}} 
&0.716  &0.638  &\multicolumn{1}{c|}{0.759}
&0.713  &0.634  &\multicolumn{1}{c|}{0.756}
&0.687  &0.603  &\multicolumn{1}{c|}{0.733}
&0.658  &0.564  &\multicolumn{1}{c}{0.712} \\
\hline
\multicolumn{1}{c|}{COTET}
&\textbf{0.754}  &\textbf{0.683}  &\multicolumn{1}{c|}{\textbf{0.794}}
&\textbf{0.748}  &\textbf{0.678}  &\multicolumn{1}{c|}{\textbf{0.785}}
&\textbf{0.731}  &\textbf{0.656}  &\multicolumn{1}{c|}{\textbf{0.773}}
&\textbf{0.707}  &\textbf{0.628}  &\multicolumn{1}{c}{\textbf{0.748}} \\
\hline
\end{tabular*}
\caption{Evaluation with different relation types dropping rates on FB15kET. Best scores are highlighted in \textbf{bold}.
}
\label{table_ablation_with_different_relation_type_dropping_rates}
\end{table*}

\smallskip \noindent {\textbf{Different  Dropping Rates of Relational Neighbors.} Relational neighbors of an entity provide supporting information for its representation. To verify the performance of COTET in the scenario where entities have different number of relational neighbors~\cite{Hu_2022}, we randomly remove  entity neighbors in the relational graph $\mathcal{G}$. The corresponding results are shown in Table~\ref{table_ablation_with_different_relational_neighbors_dropping_rates}. We note that even after removing different proportions of relational neighbors, COTET still achieves optimal performance. Specifically, when the removal ratio is 90\%, COTET's MRR  is 3.0\% higher  than that of TET (the best performing baseline). In addition,  note that as the removal ratio increases, COTET's performance does not decrease catastrophically. For example, when increasing the removal ratio from 25\% to 90\%, the MRR indicator value only drops by 4.7\%. It confirms the strong robustness of COTET to the number of relational neighbors. This  is mainly because it pays more attention to the content of type neighbors in the three view designs and does not integrate relational neighbors into the representation process of entities and types. Relational neighbors are only used in  the MPM module, so the problem of insufficient structural information that may be caused by the reduction in the number of relational neighbors is compensated by type neighbors.}

\smallskip \noindent \vic{\textbf{Different Dropping Rates of Relational Types.} We consider in this setting the deletion  of a proportion of the relational types of an entity, the corresponding results are shown in the Table~\ref{table_ablation_with_different_relation_type_dropping_rates}. We can observe that even with smaller number of relational types, COTET can still achieve the best performance under different dropping rates. Specifically, when the proportion of dropping relational types is 90\%, COTET's MRR index can reach 0.707, which can even compete with MiNer when the proportion of dropping relational types is 50\% (MiNer's MRR index value oin this case is 0.716). This  demonstrates the robustness of COTET in the presence of significant variations in the number of relational types.}

\begin{table*}[!ht]
\renewcommand\arraystretch{1.1}
\setlength{\tabcolsep}{0.795em}
\centering
\small
\begin{tabular*}{0.825\linewidth}{@{}c|cccccccccc@{}}
\hline
\multicolumn{1}{c|}{} & \multicolumn{1}{c|}{\textbf{Datasets}} & \multicolumn{4}{c|}{\textbf{FB15kET}} & \multicolumn{4}{c}{\textbf{YAGO43kET}}\\
\hline
\multicolumn{1}{c|}{} & \multicolumn{1}{c|}{\textbf{Setting}} & \multicolumn{1}{c}{\textbf{MRR}} & \multicolumn{1}{c}{\textbf{Hits@1}} & \multicolumn{1}{c}{\textbf{Hits@3}} & \multicolumn{1}{c|}{\textbf{Hits@10}} & \multicolumn{1}{c}{\textbf{MRR}} & \multicolumn{1}{c}{\textbf{Hits@1}} & \multicolumn{1}{c}{\textbf{Hits@3}} & \multicolumn{1}{c}{\textbf{Hits@10}} \\
\hline
\multicolumn{1}{c|}{\multirow{3}{*}{view}}
& \multicolumn{1}{c|}{w/o $\mathcal{T}_{e2t}$} &0.738  &0.661 &0.783  &\multicolumn{1}{c|}{0.886}  &0.516  &0.415 &0.577 &0.698 \\
& \multicolumn{1}{c|}{w/o $\mathcal{T}_{e2c}$} &0.719  &0.644 &0.758  &\multicolumn{1}{c|}{0.867}  &0.495  &0.386 &0.561 &0.697 \\
& \multicolumn{1}{c|}{w/o $\mathcal{T}_{tct}$} &0.752  &0.685 &0.787  &\multicolumn{1}{c|}{0.888}  &0.514  &0.408 &0.578 &0.710 \\
\cdashline{1-10}
\multicolumn{1}{c|}{\multirow{3}{*}{pooling}}
& \multicolumn{1}{c|}{w/o $\textbf{S}^w$} &0.730  &0.658 &0.766  &\multicolumn{1}{c|}{0.871}  &0.494  &0.393 &0.556 &0.679 \\
& \multicolumn{1}{c|}{w/o $\textbf{S}^m$} &0.759  &0.691 &0.798  &\multicolumn{1}{c|}{0.894}  &0.500  &0.399 &0.559 &0.687 \\
& \multicolumn{1}{c|}{w/o $\textbf{S}^a$} &0.736  &0.664 &0.773  &\multicolumn{1}{c|}{0.881}  &0.496  &0.393 &0.557 &0.690 \\
\cdashline{1-10}
\multicolumn{1}{c|}{\multirow{5}{*}{module}}
& \multicolumn{1}{c|}{w/ \texttt{MLP}} &0.719  &0.643 &0.758  &\multicolumn{1}{c|}{0.866}  &0.467  &0.357 &0.531 &0.670 \\
& \multicolumn{1}{c|}{w/ \texttt{ATT}} &0.719  &0.645 &0.756  &\multicolumn{1}{c|}{0.867}  &0.476  &0.367 &0.541 &0.675 \\
& \multicolumn{1}{c|}{w/o \texttt{OTA}} &0.758  &0.690 &0.796  &\multicolumn{1}{c|}{\textbf{0.895}}  &0.514  &0.407 &0.580 &0.710 \\
& \multicolumn{1}{c|}{w/o \texttt{MPM}} &0.685  &0.603 &0.726  &\multicolumn{1}{c|}{0.849}  &0.395  &0.306 &0.435 &0.574 \\
& \multicolumn{1}{c|}{w/o \texttt{BDCE}} &0.735  &0.653 &0.786  &\multicolumn{1}{c|}{0.890}  &0.511  &0.400 &0.580 &0.713 \\
\hline
\multicolumn{1}{c|}{} & \multicolumn{1}{c|}{COTET} &\textbf{0.762}  &\textbf{0.694} &\textbf{0.801}  &\multicolumn{1}{c|}{\textbf{0.895}}  &\textbf{0.529}  &\textbf{0.419} &\textbf{0.597} &\textbf{0.729} \\
\hline
\end{tabular*}
\caption{Ablation studies on the FB15kET and YAGO43kET datasets with different settings. Best scores are highlighted in \textbf{bold}.}
\label{table_ablation_result_different_graph}
\end{table*}

\begin{table*}[!ht]
\renewcommand\arraystretch{1.1}
\setlength{\tabcolsep}{0.66em}
\centering
\small
\begin{tabular*}{0.69\linewidth}{@{}cccccccccc@{}}
\hline
\multicolumn{1}{c|}{\textbf{Datasets}} & \multicolumn{4}{c|}{\textbf{FB15kET}} & \multicolumn{4}{c}{\textbf{YAGO43kET}}\\
\hline
\multicolumn{1}{c|}{\textbf{Function}} & \multicolumn{1}{c}{\textbf{MRR}} & \multicolumn{1}{c}{\textbf{Hits@1}} & \multicolumn{1}{c}{\textbf{Hits@3}} & \multicolumn{1}{c|}{\textbf{Hits@10}} & \multicolumn{1}{c}{\textbf{MRR}} & \multicolumn{1}{c}{\textbf{Hits@1}} & \multicolumn{1}{c}{\textbf{Hits@3}} & \multicolumn{1}{c}{\textbf{Hits@10}} \\
\hline
\multicolumn{1}{c|}{Cauchy} &0.752  &0.675 &0.799  &\multicolumn{1}{c|}{\textbf{0.895}}  &0.517  &0.403 &0.588 &0.724 \\
\multicolumn{1}{c|}{Gumbel} &0.750  &0.674 &0.796  &\multicolumn{1}{c|}{0.893}  &0.511  &0.397 &0.584 &0.719 \\
\multicolumn{1}{c|}{Laplace} &0.758  &0.688 &0.797  &\multicolumn{1}{c|}{0.895}  &0.522  &0.414 &0.589 &0.716 \\
\multicolumn{1}{c|}{Beta} &\textbf{0.762}  &\textbf{0.694} &\textbf{0.801}  &\multicolumn{1}{c|}{\textbf{0.895}}  &\textbf{0.529}  &\textbf{0.419} &\textbf{0.597} &\textbf{0.729} \\
\hline
\end{tabular*}
\caption{Ablation studies on the FB15kET and YAGO43kET datasets with different distribution functions in BDCE loss. Best scores are highlighted in \textbf{bold}.}
\label{table_ablation_result_different_distribution_function}
\end{table*}

\subsection{Ablation Studies}
\vic{To address \textbf{RQ2}, we conduct  several ablation studies, including different views, different pooling strategies, different modules, and different distribution functions. The corresponding results are shown in Tables~\ref{table_ablation_result_different_graph} and~\ref{table_ablation_result_different_distribution_function}.}


\smallskip \noindent \textbf{Impact of Different Views.} \vic{In Table~\ref{table_ablation_result_different_graph} we examine the individual contribution of different views. The removal of different views has varying degrees of performance drop. The entity-cluster view is shown to be the most important,  its removal leads to a significant performance drop.  Intuitively, this shows that  coarse-grained cluster information can provide  important high-level semantic content, helping to  alleviate the problem of blurred boundaries of fine-grained type prediction. On the opposite side of the spectrum, we observe that the type-cluster-type view obtains the lowest performance gain. The main reason  is that the type-cluster view is smaller and denser than the other views, resulting in a lower distinguishability of the knowledge obtained from type and cluster embeddings.  }


\smallskip \noindent \textbf{Impact of Different Pooling Strategies.} \vic{We investigate the effect of different pooling sub-modules  on the performance. Table~\ref{table_ablation_result_different_graph} shows that the removal of multi-head weight pooling has the most significant impact on performance. This is natural since weight pooling enables the model to learn independent weight coefficients for each neighbor's individual inference result, thereby distinguishing the impact of each neighbor on the final prediction. The removal of max and average pooling also lead to some degree of performance drop, confirming the importance of these two pooling methods.}


\smallskip \noindent \textbf{Impact of Different Modules.} \vic{We study  the effect of each module on the performance. The results are reported in Table~\ref{table_ablation_result_different_graph}. We  observe that the removal of any module leads to a decrease in performance. By removing the OTA module, the performance reduces 0.4\% and 1.5\% in MRR on the FB15kET and YAGO43kET datasets, respectively. \vic{In addition, we replace the OTA module with two simple and direct mechanisms: \textit{i}) the embeddings from different views of an entity or a type are concatenated and then a multi-layer perceptron is introduced for dimension transformation (shown in the w/ \texttt{MLP} row in Table 3). \textit{ii}) an attention-based method is used to fuse the  embeddings from different views of an entity or a type (shown in the w/ \texttt{ATT} row in Table 3). We observe that both alternatives cause  huge performance deficiencies, further confirming the importance of the OTA module.}
The removal of the MPM module results in a decrease of around 5.7\% and 13.4\% in MRR on the FB15kET and YAGO43kET datasets, respectively. 
This is mainly attributed to the OTA module performing alignment operations on the heterogeneous space of embeddings across views, which enables an effective fusion of view-specific  embeddings. With  the unified embeddings at hand, MPM uses different pooling mechanisms to obtain the final predictions, which integrates the prediction results of each entity neighbor.}

\smallskip \noindent \textbf{Impact of Different Distribution Functions.} \vic{By introducing the distribution-based cross-entropy loss function, we can alleviate the false negative problem during training. We conducted an ablation study on the probability distribution function that $f_{(\alpha, \beta)}$ may adopt in Equation~\ref{equation_4}. Table~\ref{table_ablation_result_different_distribution_function} presents the obtained results. Note that  any probability distribution function with a parabolic shape opening downwards between 0 and 1 can be used as a replacement for function $f_{(\alpha, \beta)}$. Here, we use Cauchy, Gumbel, and Laplace for experimental comparison. Specifically, we set the (\emph{location, scale}) hyperparameters of Cauchy, Gumbel, and Laplace to  (0.5, 1.0), (1.0, 3.0), and (0.5, 0.5), respectively. We observe that the Beta function yields the best performance, but  in general, choosing different probability distribution functions does not lead to significant performance differences. }

\begin{table*}[!ht]
\renewcommand\arraystretch{1.1}
\setlength{\tabcolsep}{0.67em}
\centering
\small
\begin{tabular*}{0.88\linewidth}{@{}c|cccccccccc@{}}
\hline
&\multicolumn{1}{c|}{\textbf{Datasets}} & \multicolumn{4}{c|}{\textbf{FB15kET}} & \multicolumn{4}{c}{\textbf{YAGO43kET}}\\
\hline
\multicolumn{1}{c|}{\textbf{Settings}}  &\multicolumn{1}{c|}{\textbf{Values}} & \multicolumn{1}{c}{\textbf{MRR}} & \multicolumn{1}{c}{\textbf{Hits@1}} & \multicolumn{1}{c}{\textbf{Hits@3}} & \multicolumn{1}{c|}{\textbf{Hits@10}} & \multicolumn{1}{c}{\textbf{MRR}} & \multicolumn{1}{c}{\textbf{Hits@1}} & \multicolumn{1}{c}{\textbf{Hits@3}} & \multicolumn{1}{c}{\textbf{Hits@10}} \\
\hline
\multicolumn{1}{c|}{\multirow{4}{*}{$\theta$}} 
&\multicolumn{1}{c|}{0.5} &0.745  &0.669 &0.786  &\multicolumn{1}{c|}{0.890}  &\textbf{0.529}  &\textbf{0.419} &\textbf{0.597} &\textbf{0.729} \\
&\multicolumn{1}{c|}{0.6} &0.750  &0.679 &0.789  &\multicolumn{1}{c|}{0.888}  &0.524  &0.417 &0.591 &0.720 \\
&\multicolumn{1}{c|}{0.7} &\textbf{0.762}  &\textbf{0.694} &\textbf{0.801}  &\multicolumn{1}{c|}{\textbf{0.895}}  &0.527  &0.417 &0.595 &0.726 \\
&\multicolumn{1}{c|}{0.8} &0.749  &0.675 &0.790 &\multicolumn{1}{c|}{0.892}  &0.520  &0.412 &0.588 &0.718 \\
\cdashline{1-10}
\multicolumn{1}{c|}{\multirow{4}{*}{$\alpha$ and $\beta$}} 
&\multicolumn{1}{c|}{1.5} &0.744  &0.670 &0.787  &\multicolumn{1}{c|}{0.889}  &0.524  &0.415 &0.594 &0.723 \\
&\multicolumn{1}{c|}{2.0} &\textbf{0.762}  &\textbf{0.694} &\textbf{0.801}  &\multicolumn{1}{c|}{\textbf{0.895}}  &\textbf{0.529}  &\textbf{0.419} &\textbf{0.597} &\textbf{0.729} \\
&\multicolumn{1}{c|}{2.5} &0.739  &0.656 &0.788  &\multicolumn{1}{c|}{\textbf{0.895}}  &0.524  &0.416 &0.591 &0.717 \\
&\multicolumn{1}{c|}{3.0} &0.719  &0.627 &0.780 &\multicolumn{1}{c|}{0.888}  &0.522  &0.414 &0.591 &0.718 \\
\cdashline{1-10}
\multicolumn{1}{c|}{\multirow{4}{*}{$h_i$}} 
&\multicolumn{1}{c|}{\{0.5, 1.0, 1.5, 2.0, 2.5\}} &\textbf{0.762}  &\textbf{0.694} &\textbf{0.801}  &\multicolumn{1}{c|}{\textbf{0.895}}  &\textbf{0.529}  &\textbf{0.419} &\textbf{0.597} &\textbf{0.729} \\
&\multicolumn{1}{c|}{\{1.0, 2.0, 3.0, 4.0, 5.0\}} &0.748  &0.679 &0.786  &\multicolumn{1}{c|}{0.888} &0.521  &0.414 &0.586 &0.714 \\
&\multicolumn{1}{c|}{\{1.5, 2.0, 2.5, 3.0, 3.5\}} &0.749  &0.677 &0.788  &\multicolumn{1}{c|}{0.889}  &0.519  &0.411 &0.587 &0.716 \\
&\multicolumn{1}{c|}{\{2.0, 2.5, 3.5, 4.5, 5.5\}} &0.751  &0.683 &0.788 &\multicolumn{1}{c|}{0.887}  &0.523  &0.417 &0.591 &0.714 \\
\hline
\end{tabular*}
\caption{The Parameter sensitivity experiments on FB15kET and YAGO43kET datasets with different $\theta$ hyperparameter in BDCE module, different $\alpha$ and $\beta$ hyperparameter in Beta function, and different temperature $h_i$ in multi-head weight pooling sub-module. Best scores are highlighted in \textbf{bold}.}
\label{table_ablation_result_appendix}
\end{table*}

\subsection{Parameter Sensitivity}
\vic{To address \textbf{RQ3}, we carry out  parameter sensitivity experiments on the FB15kET and YAGO43kET datasets, including: a) impact of $\theta$ hyperparameter; b) impact of $\alpha$ and $\beta$ hyperparameters; c) impact of temperatures of different heads. The corresponding results are shown in Table~\ref{table_ablation_result_appendix}.}

\smallskip \noindent \textbf{Impact of $\theta$ Hyperparameter.} The parameter $\theta$ in Equation~\eqref{equation_4} determines the importance of the scores of negative samples during the training process. We set the weight values of $\theta$ to $\{0.5, 0.6, 0.7, 0.8\}$ to investigate its impact on the final prediction. The corresponding results are shown in Table~\ref{table_ablation_result_appendix}. We can observe that the choice of parameter $\theta$ does indeed affect the performance, and it cannot be guaranteed that using the same value of $\theta$ will always yield the best results across different datasets. In fact, the primary role of parameter $\theta$ is to 
prevent the model from focusing too much on false negatives and on easily learnable samples, and instead  encourage it to learn from hard samples. In practice, it is necessary to adjust parameter $\theta$ based on the characteristics of a particular dataset.

\medskip \noindent \textbf{Impact of $\alpha$ and $\beta$ Hyperparameters.} The Beta function in the BDCE module involves two hyperparameters, $\alpha$ and $\beta$, which can control the shape of the Beta distribution. When $\alpha$ and $\beta$ are equal, the probability density function of Beta forms a downward-opening parabolic curve, which exactly fits the scenario required in BDCE to address the false negatives problem. The results of the ablation experiment with different values of $\alpha$ and $\beta$ are shown in Table~\ref{table_ablation_result_appendix}. We can observe that the best performance is achieved when the value is set to 2.0. Moreover, one can observe that the impact of different values on the FB15kET dataset is more pronounced than on the YAGO43kET dataset. For instance, setting $\alpha$ and $\beta$  to 2.0 instead of 3.0  leads to a 4.3\% improvement in the MRR metric for FB15kET, while YAGO43kET only shows a 0.7\% improvement. This difference may arise from a higher probability of false negatives during the negative sampling process on the FB15kET dataset. The  Beta function can alleviate this phenomenon, making the choice of Beta function parameter values more sensitive.

\medskip \noindent \textbf{Impact of Temperatures of Different Heads.} The value of parameter $h_i$ in Equation~\eqref{equation_2} is used to appropriately expand or scale down the prediction results for each neighbor.
We propose a simple way of using multi-head attention to achieve multi-faceted adjustment of multiple prediction results. For example, using five attention heads, we set four combinations of head weights. The corresponding experimental results are shown in Table~\ref{table_ablation_result_appendix}. We can observe that using different head weights does indeed have a certain impact on the model's performance, with a greater impact on FB15kET. Considering that the head temperatures can take  arbitrary values, it is not feasible to provide exhaustive experimental results for all temperature coefficients. Therefore, setting the temperature coefficients needs to be adapted based on experience and the dataset.


\subsection{Complexity Analysis}
\vic{\vic{To address \textbf{RQ4},} we analyze the COTET complexity on the FB15kET and YAGO43kET datasets from two perspectives: time complexity and space complexity. The corresponding results are shown in Table~\ref{table_complexity_analysis}. The time complexity refers to the number of floating-point operations (\#FLOPs) required during training or inference, while the space complexity can be measured by the amount of model's parameters (\#Params). The larger the values of \#FLOPs and \#Params respectively indicate that training the model requires more computing time and greater graphics memory overhead. We can observe that when compared with CET and MiNer, COTET can significantly reduce the computational cost, i.e., \#FLOPs. Specifically, compared with MiNer, on the FB15kET dataset, COTET's FLOPs are only 14.15\% of MiNer's. Although COTET has more parameters than MiNer, this overhead is acceptable in the current context of increasing GPU memory size. In fact, on FB15kET dataset, COTET only occupies 2.703G video memory space.}

\begin{table}[!hp]
\renewcommand\arraystretch{1.2}
\setlength{\tabcolsep}{0.22cm}
\centering
\small
\begin{tabular*}{\linewidth}{@{}ccccc@{}}
\toprule
\multicolumn{1}{c|}{} & \multicolumn{2}{c|}{\textbf{FB15kET}} & \multicolumn{2}{c}{\textbf{YAGO43kET}}\\
\midrule
\multicolumn{1}{c|}{\textbf{Model}} &\multicolumn{1}{c}{\textbf{\#FLOPs}} &\multicolumn{1}{c|}{\textbf{\#Params}}&\multicolumn{1}{c}{\textbf{\#FLOPs}} &\multicolumn{1}{c}{\textbf{\#Params}}\\
\midrule
\multicolumn{1}{c|}{CompGCN} &279.315M    &\multicolumn{1}{c|}{392.284K}   &645.330M   &4.594M \\
\multicolumn{1}{c|}{RGCN}    &45.875M    &\multicolumn{1}{c|}{361.984K}   &578.330M   &4.563M \\
\multicolumn{1}{c|}{CET}     &3.423G   &\multicolumn{1}{c|}{361.984K}   &14.820G     &4.563M \\
\multicolumn{1}{c|}{TET}     &2.016G     &\multicolumn{1}{c|}{911.104K}   &8.406G     &5.113M \\
\multicolumn{1}{c|}{MiNer}   &6.613G     &\multicolumn{1}{c|}{361.984K}   &34.739G    &4.563M \\
\multicolumn{1}{c|}{{COTET}} &936.496M &\multicolumn{1}{c|}{720.984K}      &11.603G    &9.082M \\
\bottomrule
\end{tabular*}
\caption{The amount of calculations and parameters required by different models on different datasets. The unit abbreviations are: Billions (G), Millions (M), Kilo (K).}
\label{table_complexity_analysis}
\end{table}

\section{Conclusions and Future Work}
\vic{\vic{In this paper, we propose $\rm COTET$, an approach to KGET  which introduces coarse-grained cluster information into the  type inference. We construct entity-type, entity-cluster and type-cluster-type views with different granularities, and use an optimal transport mechanism to achieve cross-view cooperative interaction. We also design a mixture pooling strategy to aggregate prediction scores from different neighbors, and propose a distribution-based loss function to alleviate the false negative problem during training. Extensive experiments on the full, hard and easy, and sparse neighbor-connection versions subsets show COTET's strong performance. It improves SoTA by considerable margins. In the next step, we will consider following two aspects researches: on the one hand, $\rm COTET$ (and all existing works) focuses on the so-call transductive setting, i.e., the inferring missing types that appear in the prediction stage must be present in the training phase, for those types that do not appear in the training stage (inductive setting scenario), reasonable inferences cannot be made. As next step, we consider tackling the more challenging inductive KGET task. On the other hand, entities in the FB15k and YAGO datasets contain textual description knowledge. How to reasonably combine this  knowledge and structural knowledge to improve performance is worth exploring.}}

\IEEEdisplaynontitleabstractindextext

%
\IEEEpeerreviewmaketitle

\bibliographystyle{IEEEtran}
\bibliography{tkde}

\end{document}